\documentclass[10pt,twocolumn,letterpaper]{article}

\usepackage{iccv}
\usepackage{times}
\usepackage{epsfig}
\usepackage{graphicx}
\usepackage{amsmath}
\usepackage{amssymb}

\usepackage{booktabs}
\usepackage{float}

\usepackage{xcolor}

\usepackage{enumitem}
\setlist[itemize]{noitemsep, nolistsep,leftmargin=*}

\newcommand{\SE}{\mathrm{SE}}
\newcommand{\SO}{\mathrm{SO}}

\newcommand{\R}{\mathbb{R}}
\newcommand{\bx}{\boldsymbol{x}}
\newcommand{\by}{\boldsymbol{y}}
\newcommand{\bt}{\boldsymbol{t}}
\newcommand{\bR}{\boldsymbol{R}}
\newcommand{\bH}{\boldsymbol{H}}
\newcommand{\bS}{\boldsymbol{S}}
\newcommand{\bV}{\boldsymbol{V}}
\newcommand{\bU}{\boldsymbol{U}}

\newcommand{\XY}{\mathcal{XY}}
\newcommand{\X}{\mathcal{X}}
\newcommand{\Y}{\mathcal{Y}}
\newcommand{\F}{\mathcal{F}}
\newcommand{\N}{\mathcal{N}}

\DeclareMathOperator*{\argmin}{arg\,min}

\usepackage[pagebackref=true,breaklinks=true,letterpaper=true,colorlinks,bookmarks=false]{hyperref}
\usepackage{microtype}

\iccvfinalcopy 

\def\iccvPaperID{****} 

\begin{document}

\title{Deep Closest Point: Learning Representations for Point Cloud Registration}

\author{Yue Wang\\
Massachusetts Institute of Technology\\
77 Massachusetts Ave, Cambridge, MA 02139\\
{\tt\small yuewangx@mit.edu}
\and
Justin M. Solomon\\
Massachusetts Institute of Technology\\
77 Massachusetts Ave, Cambridge, MA 02139\\
{\tt\small jsolomon@mit.edu}
}

\maketitle

\begin{abstract}
Point cloud registration is a key problem for computer vision applied to robotics, medical imaging, and other applications. This problem involves finding a rigid transformation from one point cloud into another so that they align. Iterative Closest Point (ICP) and its variants provide simple and easily-implemented iterative methods for this task, but these algorithms can converge to spurious local optima. To address local optima and other difficulties in the ICP pipeline, we propose a learning-based method, titled Deep Closest Point (DCP), inspired by recent techniques in computer vision and natural language processing. Our model consists of three parts: a point cloud embedding network, an attention-based module combined with a pointer generation layer, to approximate combinatorial matching, and a differentiable singular value decomposition (SVD) layer to extract the final rigid transformation. We train our model end-to-end on the ModelNet40 dataset and show in several settings that it performs better than ICP, its variants (e.g., Go-ICP, FGR), and the recently-proposed learning-based method PointNetLK.  Beyond providing a state-of-the-art registration technique, we evaluate the suitability of our learned features transferred to unseen objects. We also provide preliminary analysis of our learned model to help understand whether domain-specific and/or global features facilitate rigid registration.
\end{abstract}
\section{Introduction}
\vskip 0.1in
\begin{figure}[t!]  \label{fig:guitar_human}
  \centering
 \includegraphics[width=1.0\columnwidth]{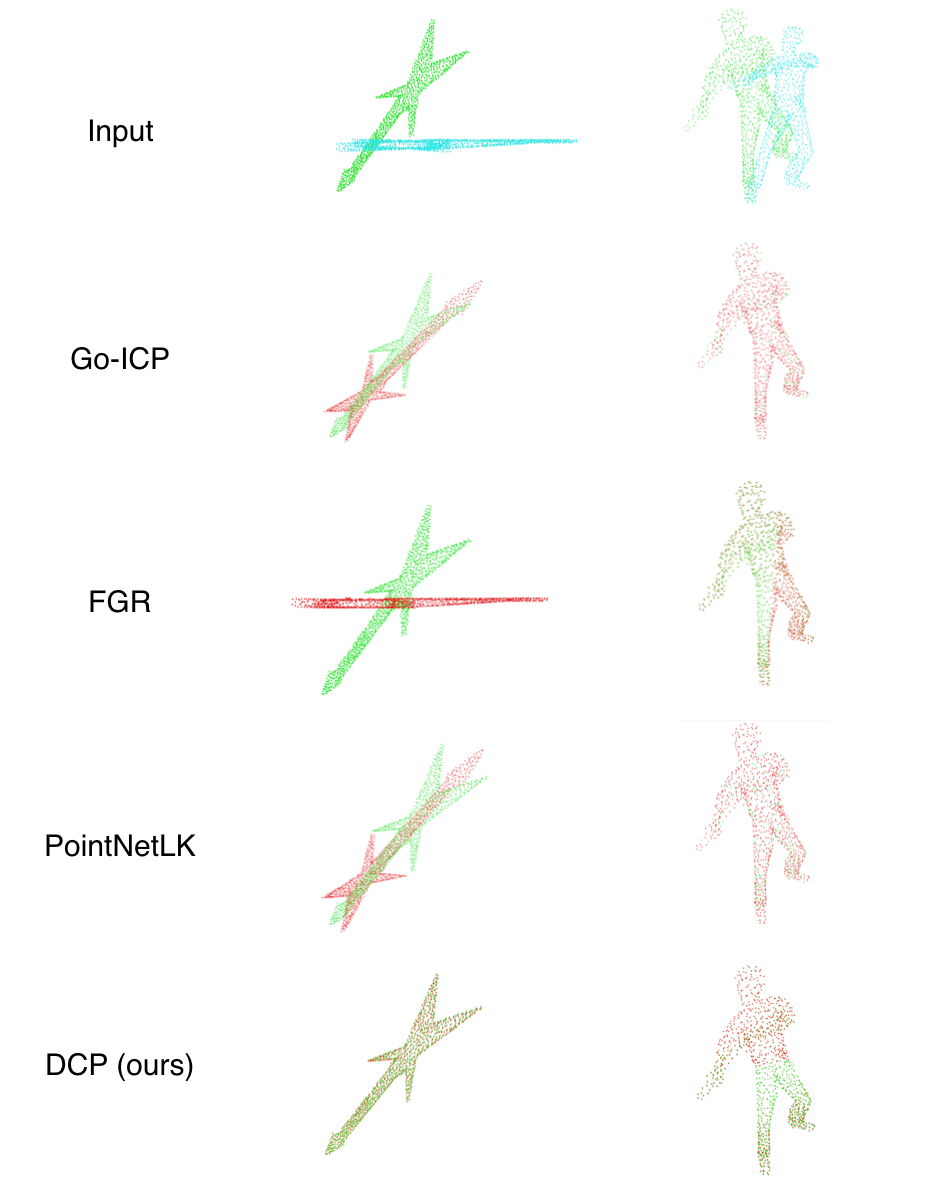}
  \caption{\textbf{Left}: a moved guitar. \textbf{Right}: rotated human. All methods work well with small transformation. However, only our method achieve satisfying alignment for objects with sharp features and large transformation.}
\vskip 0.1in
\end{figure}

Geometric registration is a key task in many computational fields, including medical imaging, robotics, autonomous driving, and computational chemistry. 
In its most basic incarnation, registration involves the prediction of a rigid motion to align one shape to another, potentially obfuscated by noise and partiality.

Many modeling and computational challenges hamper the design of a stable and efficient registration method. 
Given exact correspondences, singular value decomposition yields the globally optimal alignment; similarly, computing matchings becomes easier given some global alignment information. Given these two observations, most algorithms alternate between these two steps to try to obtain a better
result. The resultant iterative optimization algorithms, however, are prone to local optima. 

The most popular example, Iterative Closest Point (ICP) \cite{Besl1992, Segal2009}, alternates between estimating the rigid motion based on a fixed correspondence estimate and updating the correspondences to their closest matches. Although ICP monotonically decreases a certain objective function measuring alignment, due to the non-convexity of the problem, ICP often stalls in suboptimal local minima. Many methods \cite{Rusinkiewicz2001, Fitzgibbon01c, Yang2016GoICP} attempt to alleviate this issue by using heuristics to improve the matching or by searching larger parts of the motion space $\SE(3)$. These algorithms are typically slower than ICP and still do not always provide acceptable output.

In this work, we revisit ICP from a deep learning perspective, addressing key issues in each part of the ICP pipeline using modern machine learning, computer vision, and natural language processing tools. We call our resulting algorithm \emph{Deep Closest Point (DCP)}, a learning-based method that takes two point clouds and predicts a rigid transformation aligning them.

Our model consists of three parts: (1) We map the input point clouds to permutation/rigid-invariant embeddings that help identify matching pairs of points (we compare PointNet \cite{QiSMG17} and DGCNN \cite{dgcnn} for this step); then, (2) an attention based module combining pointer network \cite{Vinyals2015pn,Vaswani2017} predicts a soft matching between the point clouds;  
and finally, (3) a differentiable singular value decomposition layer predicts the rigid transformation. We train and test our model end-to-end on ModelNet40 \cite{Wu2015modelnet} in various settings, showing our model is not only efficient but also outperforms ICP and its extensions, as well as the recently-proposed PointNetLK method \cite{Goforth2019}. 
Our learned features generalize to unseen data, suggesting that our model is learning salient geometric features. 

\paragraph*{Contributions:} 
Our contributions include the following:
    \begin{itemize} 
    \setlength\itemsep{0em}
    \item We identify sub-network architectures designed to address difficulties in the classical ICP pipeline. 
    \item We propose a simple architecture to predict a rigid transformation aligning two point clouds. 
    \item We evaluate efficiency and performance in several settings and provide an ablation study to support details of our construction.  
    \item We analyze whether local or global features are more useful for registration. 
    \item We release our code\footnote{\url{https://github.com/WangYueFt/dcp}}, to facilitate reproducibility and future research.
    \end{itemize}
\section{Related Work}
\begin{figure*}[t!]  \label{fig:network}
  \vskip 0.1in
  \centering
  \begin{tabular}{@{}cc@{}}
 \includegraphics[height=1.6in]{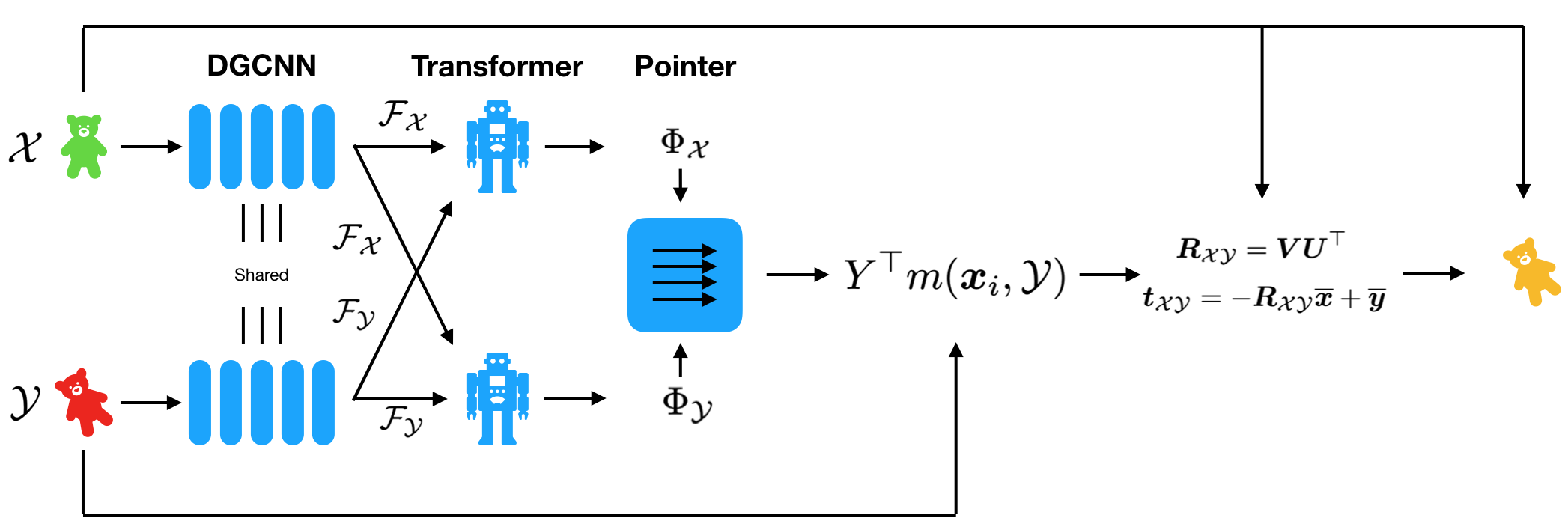}&
  \includegraphics[height=1.6in]{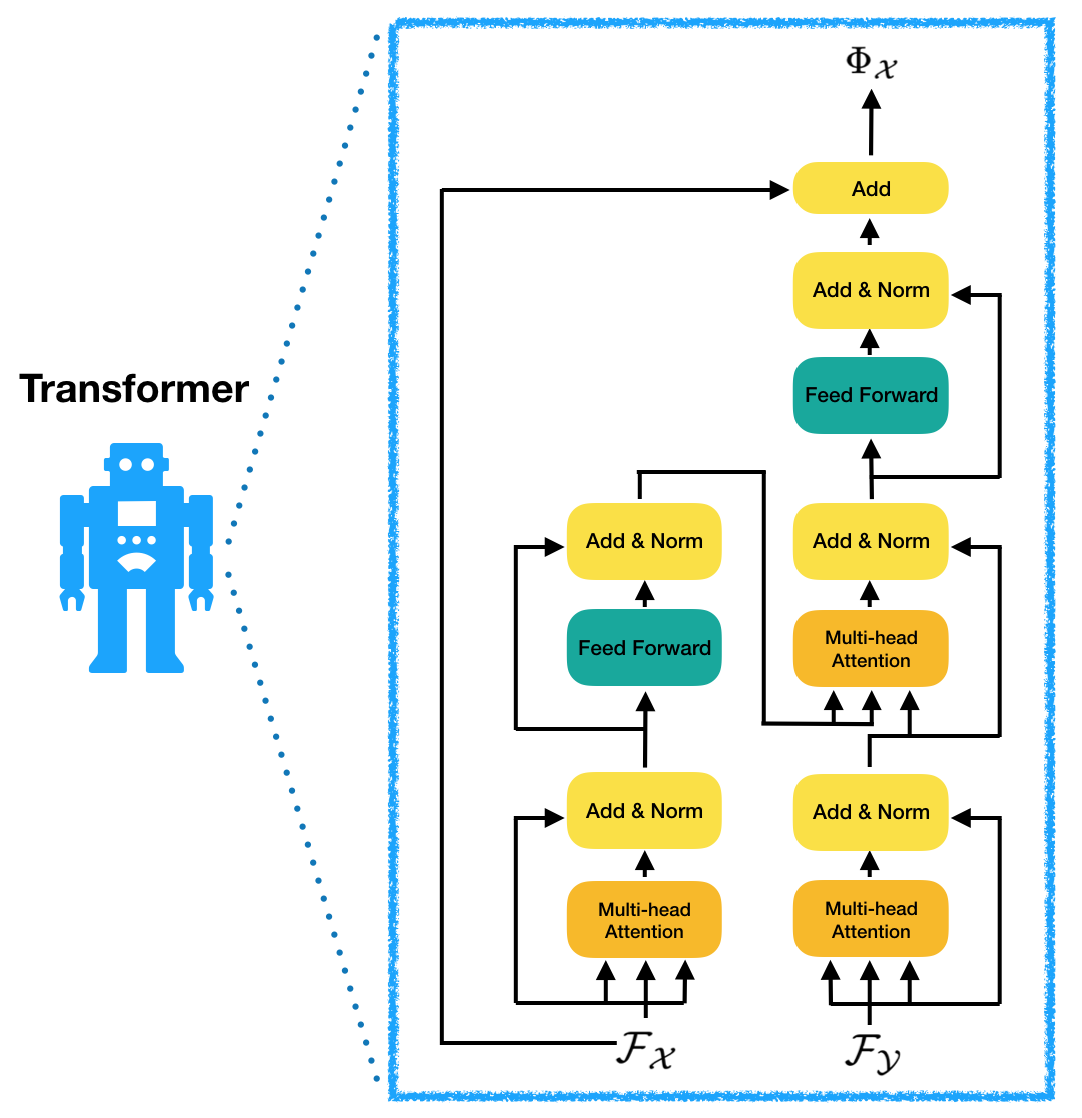} \\
  (a) Network architecture & (b) Transformer module
  \end{tabular}
  \vspace{.1in}
  \caption{Network architecture for DCP, including the Transformer module for DCP-v2.}
  \vskip 0.1in
\end{figure*}

\paragraph*{Traditional point cloud registration methods:} ICP \cite{Besl1992} is the best-known algorithm for solving rigid registration problems; it alternates between finding point cloud correspondences and solving a least-squares problem to update the alignment. 
ICP variants \cite{Rusinkiewicz2001, Segal2009, Bouaziz2013sgp} consider issues with the basic method, like noise, partiality, and sparsity; probabilistic models \cite{agamennoniIROS16, iser_2016_hinzmann, Hahnel02probabilisticmatching} also can improve resilience to uncertain data. ICP can be viewed as an optimization algorithm searching jointly for a matching and a rigid alignment; hence, \cite{Fitzgibbon01c} propose using the Levenberg--Marquardt algorithm to  optimize the objective directly, which can yield a better solution. 
For more information, \cite{Pomerleau2015, Rusinkiewicz2001} summarize  ICP and its variants developed over the last 20 years. 

ICP-style methods are prone to local minima due to non-convexity. To find a good optimum in polynomial time, Go-ICP \cite{Yang2016GoICP} uses a branch-and-bound (BnB) method to search the motion space $\SE(3)$. It outperforms local ICP methods when a global solution is desired but is several orders of magnitude slower than other ICP variants despite using local ICP to accelerate the search process. 
Other methods attempt to identify global optima using Riemannian optimization \cite{Rosen16wafr-sesync}, convex relaxation \cite{Maron2016}, and mixed-integer programming \cite{Izatt2017}.

\paragraph*{Learning on graphs and point sets:} A broad class of deep architectures for geometric data termed \textit{geometric deep learning} \cite{bronstein2017geometric} includes recent methods learning on graphs \cite{Zonghan2019, Zhou2018GraphNN, Fey2018splinecnn} and point clouds \cite{QiSMG17, Qi2017pn++, dgcnn, Zaheer2017}.

The \emph{graph neural network} (GNN) is introduced in \cite{Scarselli2009}; similarly, \cite{Duvenaud2015} defines convolution on graphs (GCN) for molecular data. \cite{kipf2016semi} uses renormalization to adapt to graph structure and applies GCN to semi-supervised learning on graphs. MoNet \cite{Monti17} learns a dynamic aggregation function based on graph structure, generalizing GNNs. Finally, graph attention networks (GATs) \cite{velickovic2018graph} incorporate multi-head attention into GCNs. DGCNN \cite{dgcnn} can be regarded as graph neural network applied to point clouds with dynamic edges.

Another branch of geometric deep learning includes PointNet \cite{QiSMG17} and other algorithms designed to process point clouds. PointNet can be seen as applying GCN to graphs without edges, mapping points in $\R^3$ to high-dimensional space. PointNet only encodes global features gathered from the point cloud's embedding, 
impeding application to tasks involving local geometry. To address this issue, PointNet++ \cite{Qi2017pn++} applies a shared PointNet to $k$-nearest neighbor clusters to learn local features. As an alternative, DGCNN \cite{dgcnn} explicitly recovers the graph structure in both Euclidean space and feature space and applies graph neural networks to the result. PCNN \cite{Atzmon2018} uses an extension operator to define convolution on point clouds, while PointCNN \cite{Li2018pointcnn} applies Euclidean convolution after applying a learned transformation. Finally, SPLATNet \cite{su18splatnet} encodes point clouds on a lattice and performs bilateral convolution. All these works aim to apply convolution-like operations to point clouds and extract local geometric features.  
 
\paragraph*{Sequence-to-sequence learning and pointer networks:} Many tasks in natural language processing, including machine translation, language modeling, and question answering, can be formulated as sequence-to-sequence problems (seq2seq). \cite{Sutskever2014} first uses deep neural networks (DNN) to address seq2seq problems at large scale. Seq2seq, however, often involves predicting discrete tokens corresponding to positions in the input sequence. This problem is difficult because there is an exponential number of possible matchings between input and output positions. 
Similar problems can be found in optimal transportation \cite{Justin2018ot, Gabriel2019}, combinatorial optimization \cite{Ivanov2002}, and graph matching \cite{Yan2016}. To address this issue, 
In our registration pipeline, we use a related method to Pointer Networks \cite{Vinyals2015}, which use attention as a pointer to select from the input sequence. In each output step, a Pointer Network predicts a distribution over positions and uses it as a ``soft pointer.'' The pointer module is fully differentiable, and the whole network can be trained end-to-end.  

\paragraph*{Non-local approaches:} To denoise images, non-local means \cite{Buades2005} leverages the simple observation that Gaussian noise can be removed by non-locally weighted averaging all pixels in an image. 
Recently, non-local neural networks \cite{NonLocal2018} have been proposed to capture long-range dependencies in video understanding; \cite{Xie2018denoise} uses the non-local module to denoise feature maps to defend against adversarial attacks. Another instantiation of non-local neural networks, known as \emph{relational networks} \cite{Santoro2017}, has shown effectiveness in visual reasoning \cite{Santoro2017}, meta-learning \cite{sung2018learning}, object detection \cite{hu2017relation}, and reinforcement learning \cite{Flores2018}. Its counterpart in natural language processing, attention, is arguably the most fruitful recent advance in this discipline. \cite {Vaswani2017} replaces recurrent neural networks \cite{Jordan1990, Hochreiter1997} with a model called Transformer, consisting of several stacked multi-head attention modules. Transformer-based models \cite{Devlin2018, radford2019language} outperform other recurrent models by a considerable amount in natural language processing. In our work, we also use Transformer to learn contextual information of point clouds. 

\section{Problem Statement}

In this section, we formulate the rigid alignment problem and discuss the ICP algorithm, highlighting key issues in the ICP pipeline. 
 We use $\X$ and $\Y$ to denote two point clouds, 
 where $\X=\{\bx_1, \ldots, \bx_i, \ldots, \bx_N\}\subset\R^3$ and $\Y=\{\by_1, \ldots, \by_j, \ldots, \by_M\}\subset\R^3$. 
 For ease of notation, we consider the simplest case, in which $M=N$; the methods we describe here extend easily to the $M\neq N$ case.
 
 In the rigid alignment problem, we assume $\Y$ is transformed from $\X$ by an unknown rigid motion. We denote the rigid transformation as $[\bR_\XY, \bt_\XY]$ where $\bR_\XY \in \SO(3)$ and $\bt_\XY \in \R^3$. We want to minimize the mean-squared error $E(\bR_\XY, \bt_\XY)$, which---if $\X$ and $\Y$ are ordered the same way (meaning $\bx_i$ and $\by_i$ are paired)---can be written
 \begin{equation} \label{icp:paired:objective}
     \begin{split}
         E(\bR_\XY, \bt_\XY) = \frac{1}{N} \sum_i^N\lVert \bR_\XY\bx_i+\bt_\XY-\by_i  \rVert^2.
     \end{split}
 \end{equation}
 Define centroids of $\X$ and $\Y$ as
 \begin{equation} \label{icp:paired:centroid}
        \overline\bx = \frac{1}{N}\sum_{i=1}^N\bx_i \qquad\textrm{and}\qquad
        \overline\by = \frac{1}{N}\sum_{i=1}^N\by_i. 
 \end{equation}
 Then the cross-covariance matrix $\bH$ is given by
 \begin{equation} \label{icp:paired:cross_covariance}
         \bH = \sum_{i=1}^N(\bx_i-\overline\bx)(\by_i-\overline\by)^\top.
 \end{equation}
We can use the singular value decomposition (SVD) to decompose $\bH=\bU\bS\bV^\top$. Then, the alignment minimizing $E(\cdot,\cdot)$ in \eqref{icp:paired:objective} is given in closed-form by 
 \begin{equation} \label{icp:paired:close_form}
        \bR_\XY = \bV\bU^\top\qquad\textrm{and}\qquad
        \bt_\XY = -\bR_\XY\overline\bx+\overline\by.
 \end{equation}
Here, we take the convention that $\bU,\bV\in\SO(3)$, while $\bS$ is diagonal but potentially signed; this accounts for orientation-reversing choices of $\bH$. This classic orthogonal Procrustes problem assumes that the point sets are matched to each other, that is, that $\bx_i$ should be mapped to $\by_i$ in the final alignment for all $i$.  If the correspondence is unknown, however, the objective function $E$ must be revised to account for matching:
  \begin{equation} \label{icp:objective}
     \begin{split}
         E(\bR_\XY, \bt_\XY)
        \!=\!\frac{1}{N}\!\sum_i^N\lVert \bR_\XY\bx_i+\bt_\XY-\by_{m(x_i)}  \rVert^2.
     \end{split}
 \end{equation}
 Here, a mapping $m$ from each point in $\X$ to its corresponding point in $\Y$ is given by
 \begin{equation} \label{icp:mapping}
    \begin{split}
        m(\bx_i, \Y) = \argmin_j\ \lVert \bR_\XY\bx_i+\bt_\XY - \by_j \rVert
    \end{split}
 \end{equation}
 Equations \eqref{icp:objective} and \eqref{icp:mapping} form a classic \emph{chicken-and-egg} problem. If we know the optimal rigid transformation $[\bR_\XY,\bt_\XY]$, then the mapping $m$ can be recovered from \eqref{icp:mapping}; conversely, given the optimal mapping $m$, the transformation can be computed using \eqref{icp:paired:close_form}. 
 
 ICP iteratively approaches a stationary point of $E$ in \eqref{icp:objective}, including the mapping $m(\cdot)$ as one of the variables in the optimization problem. It alternates between two steps: finding the current optimal transformation based on a previous mapping $m^{k-1}$ and finding an optimal mapping $m^k$ based on the current transformation using~\eqref{icp:mapping}, where $k$ denotes the current iteration. 
 The algorithm terminates when a fixed point or stall criterion is reached. This procedure is easy to implement and relatively efficient, but it is extremely prone to local optima; a distant initial alignment yields a poor estimate of the mapping $m$, quickly leading to a situation where the algorithm gets stuck. Our goal is to use learned embeddings to recover a better matching $m(\cdot)$ and use that to compute a rigid transformation, which we will detail in next section. 

\section{Deep Closest Point}

Having established preliminaries about the rigid alignment problem, we are now equipped to present our Deep Closest Point architecture, illustrated in Figure \ref{fig:network}. In short, we embed point clouds into high-dimensional space using PointNet \cite{QiSMG17} or DGCNN \cite{dgcnn} (\S\ref{sec:dgcnn}), encode contextual information using an attention-based module (\S\ref{sec:attention}), and finally estimate an alignment using a differentiable SVD layer (\S\ref{sec:svd}). 

\subsection{Initial Features}
\label{sec:dgcnn}

The first stage of our pipeline embeds the unaligned input point clouds $\X$ and $\Y$ into a common space used to find matching pairs of points between the two clouds.  The goal is to find an embedding that quotients out rigid motion while remaining sensitive to relevant features for rigid matching.  
%
We evaluate two possible choices of learnable embedding modules, PointNet \cite{QiSMG17} and DGCNN \cite{dgcnn}. 

Since we use per-point embeddings of the two input point clouds to generate a mapping $m$ and recover the rigid transformation, we seek a feature \emph{per point} in the input point clouds rather than one feature per cloud.  For this reason, in these two network architectures, we use the representations generated before the last aggregation function, notated $\F_\X=\{\bx_1^L, \bx_2^L, ..., \bx_i^L, ..., \bx_N^L\}$ and $\F_\Y=\{\by_1^L, \by_2^L, ..., \by_i^L, ..., \by_N^L\}$, assuming a total of $L$ layers. 

In more detail, PointNet takes a set of points, embeds each by a nonlinear function from $\R^3$ into a higher-dimensional space, and optionally outputs a global feature vector for the whole point cloud after applying a channel-wise aggregation function $f$ (e.g., $\max$ or $\sum$).  
%
Let $\bx_i^l$ be the embedding of point $i$ in the $l$-th layer, and let $h_\theta^l$ be a nonlinear function in the $l$-th layer parameterized by a shared multilayer perceptron (MLP). Then, the forward mechanism is given by 
        $\bx_i^l = h_\theta^l(\bx_i^{l-1}).$ 

While PointNet largely extracts information based on the embedding of each point in the point cloud independently, 
DGCNN explicitly incorporates local geometry into its representation. In particular, given a set of points $\X$, DGCNN constructs a $k$-NN graph $\mathcal{G}$, applies a nonlinearity to the values at edge endpoints to obtain edgewise values, and performs vertex-wise aggregation ($\max$ or $\sum$) in each layer. The forward mechanism of DGCNN is thus 
\begin{equation} \label{dgcnn:forward}
    \begin{split}
        \bx_i^l = f(\{ h_\theta^l(\bx_i^{l-1}, \bx_j^{l-1}) \; \forall j \in \mathcal{N}_i\}),
    \end{split}
\end{equation}
where $\mathcal N_i$ denotes the neighbors of vertex $i$ in graph $\mathcal G$. 
While PointNet features do not incorporate local neighborhood information, we find empirically that DGCNN's local features are critical for high-quality matching in subsequent steps of our pipeline (see \S\ref{sec:ablation:PNorDGCNN}). 

\subsection{Attention}
\label{sec:attention}

Our transition from PointNet to DGCNN is motivated by the observation that the most useful features for rigid alignment are learned jointly from local and global information.  We additionally can improve our features for matching by making them \emph{task-specific}, that is, changing the features depending  on the particularities of $\X$ and $\Y$ together rather than embedding $\X$ and $\Y$ independently.  That is, the task of rigidly aligning, say, organic shapes might require different features than those for aligning mechanical parts with sharp edges. 
Inspired by the recent success of BERT \cite{Devlin2018}, non-local neural networks \cite{NonLocal2018}, and relational networks \cite{Santoro2017} using attention-based models, we design a module to learn co-contextual information by capturing \emph{self-attention} and \emph{conditional attention}.

Take $\F_\X$ and $\F_\Y$ to be the embeddings generated by the modules in \S\ref{sec:dgcnn}; these embeddings are computed independently of one another.  Our attention model learns a function $\phi:\R^{N\times P}\times\R^{N\times P}\to\R^{N\times P}$, where $P$ is embedding dimension, that provides new embeddings of the point clouds as
\begin{equation} \label{attention:residual}
    \begin{aligned}
        \Phi_\X &= \F_\X + \phi(\F_\X, \F_\Y) \\
        \Phi_\Y &= \F_\Y + \phi(\F_\Y, \F_\X)
    \end{aligned}    
\end{equation}
Notice we treat $\phi$ as a \emph{residual} term, providing an additive change to $\F_\X$ and $\F_\Y$ depending on the order of its inputs.  The idea here is that the map $\F_\X\mapsto\Phi_\X$ modifies the features associated to the points in $\X$ in a fashion that is knowledgeable about the structure of $\Y$; the map $\F_\Y\mapsto\Phi_\Y$ serves a symmetric role.  
We choose $\phi$ as an asymmetric function given by a Transformer \cite{Vaswani2017},\footnote{For details, see  \url{http://nlp.seas.harvard.edu/2018/04/03/attention.html}.} since the matching problem we encounter in rigid alignment is analogous to the sequence-to-sequence problem that inspired its development, other than their use of positional embeddings to describe where words are in a sentence. 


\subsection{Pointer Generation}
\label{sec:pointer}

The most common failure mode of ICP occurs when the matching estimate $m^k$ is far from optimal.  When this occurs, the rigid motion subsequently estimated using \eqref{icp:mapping} does not significantly improve alignment, leading to a spurious local optimum.  As an alternative, our learned embeddings are trained specifically to expose matching pairs of points using a simple procedure explained below.  We term this step \emph{pointer generation}, again inspired by terminology in the attention literature introduced in \S\ref{sec:attention}.


To avoid choosing non-differentiable hard assignments, we use a probabilistic approach that generates a (singly-stochastic) ``soft map'' from one point cloud into the other.  That is, each $\bx_i\in\X$ is assigned a probability vector over elements of $\Y$ given by
\begin{equation} \label{pointer:softmax}
    \begin{split}
        m(\bx_i, \Y) = \mathrm{softmax}(\Phi_\Y\Phi_{\bx_i}^\top).
    \end{split}
\end{equation}
Here, $\Phi_\Y\in\R^{N\times P}$ denotes the embedding of $\Y$ generated by the attention module, and $\Phi_{\bx_i}$ denotes the $i$-th row of the matrix $\Phi_\X$ from the attention module.  We can think of $m(\bx_i, \Y)$ as a \emph{soft pointer} from each $\bx_i$ into the elements of $\Y$.


\subsection{SVD Module} 
\label{sec:svd}

The final module in our architecture extracts the rigid motion from the soft matching computed in \S\ref{sec:pointer}.  We use the soft pointers to generate a matching averaged point in $\Y$ for each point in $\X$:
\begin{equation}
    \hat\by_i = Y^\top m(\bx_i,\Y) \in\R^3.
\end{equation}
Here, we define $Y\in\R^{N\times3}$ to be a matrix containing the points in $Y$.  Then, $\bR_\XY$ and $\bt_\XY$ are extracted using \eqref{icp:paired:close_form} based on the pairing $\bx_i\mapsto\hat\by_i$ over all $i$. 

To backpropagate gradients through the networks, we need to differentiate the SVD.  \cite{papadopoulo2000estimating} describes a standard means of computing this derivative; version of this calculations are included in PyTorch \cite{paszke2017automatic} and TensorFlow \cite{tensorflow2015-whitepaper}. Note we need to solve only $3\times3$ eigenproblems, small enough to be solved using simple algorithms or even (in principle) a closed-form formula.

\subsection{Loss}
\label{sec:loss}

Combined, the modules above map from a pair of point clouds $\X$ and $\Y$ to a rigid motion $[\bR_\XY,\bt_\XY]$ that aligns them to each other.  The initial feature module (\S\ref{sec:dgcnn}) and the attention module (\S\ref{sec:attention}) are both parameterized by a set of neural network weights, which must be learned during a training phase.  We employ a fairly straightforward strategy for training, measuring deviation of $[\bR_\XY,\bt_\XY]$ from ground truth for synthetically-generated pairs of point clouds.

We use the following loss function to measure our model's agreement to the ground-truth rigid motions:
\begin{equation} \label{dcn:loss}
    \begin{split}
        \mathrm{Loss} = \lVert \bR^\top_{\XY}\bR^g_{\XY} - I\rVert^2 +\lVert \bt_{\XY}-\bt^g_{\XY}  \rVert^2 + \lambda\lVert\theta\rVert^2\\
    \end{split}
\end{equation}
Here, $g$ denotes ground-truth.  The first two terms define a simple distance on $\SE(3)$.  The third term denotes Tikhonov regularization of the DCP parameters $\theta$, which serves to reduce the complexity of the network.


\section{Experiments}

\begin{figure}[t!]  
  \centering
 \includegraphics[width=1.1\columnwidth]{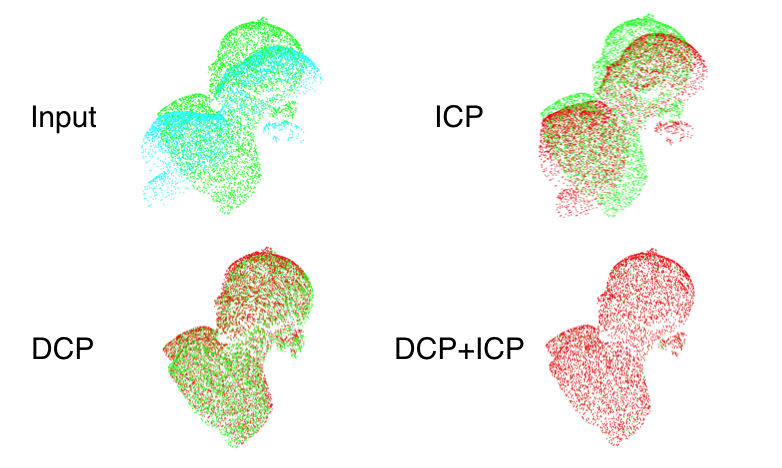}
  \caption{\textbf{Top left}: input. \textbf{Top right}: result of ICP with random initialization. \textbf{Bottom left}: initial transformation provided by DCP. \textbf{Bottom right}: result of ICP initialized with DCP. Using a good initial transformation provided by DCP, ICP converges to the global optimum.}
  \label{fig:bimba}
\end{figure}

We compare our models to ICP, Go-ICP \cite{Yang2016GoICP}, Fast Global Registration (FGR) \cite{Zhou2016fgr}, and the recently-proposed PointNetLK deep learning method \cite{Goforth2019}. We denote our model without attention (\S\ref{sec:attention}) as \textbf{DCP-v1} and the full model with attention as \textbf{DCP-v2}. Go-ICP is ported from the authors' released code. For ICP and FGR, we use the implementations in Intel Open3D \cite{Zhou2018open3d}. For PointNetLK, we adapt the code partially released by the authors.\footnote{\url{https://github.com/hmgoforth/PointNetLK}} Notice that FGR \cite{Zhou2016fgr} uses additional geometric features.

The architecture of DCP is shown in Figure \ref{fig:network}. We use 5 \textit{EdgeConv} (denoted as DGCNN \cite{dgcnn}) layers for both DCP-v1 and DCP-v2. The numbers of filters in each layer are $[64, 64, 128, 256, 512]$. In the Transformer layer, the number of heads in multi-head attention is 4 and the embedding dimension is 1024. We use LayerNorm \cite{layernorm} without Dropout \cite{dropout}. Adam \cite{KingmaB14} is used to optimize the network parameters, with initial learning rate 0.001. We divide the learning rate by 10 at epochs 75, 150, and 200, training for a total of 250 epochs. DCP-v1 does not use the Transformer module but rather employs identity mappings $\Phi_\X=\F_\X$ and $\Phi_\Y=\F_\Y$. 

We experiment on the ModelNet40 \cite{Wu2015modelnet} dataset, which consists of 12,311 meshed CAD models from 40 categories. Of these, we use 9,843 models for training and 2,468 models for testing. We follow the experimental settings of PointNet \cite{QiSMG17}, uniformly sampling 
1,024 points from each model's outer surface.  As in previous work, points are centered and rescaled to fit in the unit sphere, and no features other than $(x,y,z)$ coordinates appear in the input. 

We measure mean squared error (MSE), root mean squared error (RMSE), and mean absolute error (MAE) between ground truth values and predicted values. Ideally, all of these error metrics should be zero if the rigid alignment is perfect. All angular measurements in our results are in units of degrees. 

\subsection{ModelNet40:  Full Dataset Train \& Test}
\label{sec:full:modelnet40}

In our first experiment, we randomly divide all the point clouds in the ModelNet40 dataset into training and test sets, with no knowledge of the category label; different point clouds are used during training and during testing.   During training, we sample a point cloud $\X$. Along each axis, we randomly draw a rigid transformation; 
the rotation along each axis is uniformly sampled in $[0, 45^\circ]$ and translation is in $[-0.5, 0.5]$. $\X$ and a transformation of $\X$ by the rigid motion are used as input to the network, which is evaluated against the known ground truth using \eqref{dcn:loss}.

\begin{figure*}[t!] 
  \centering
 \includegraphics[width=2.0\columnwidth]{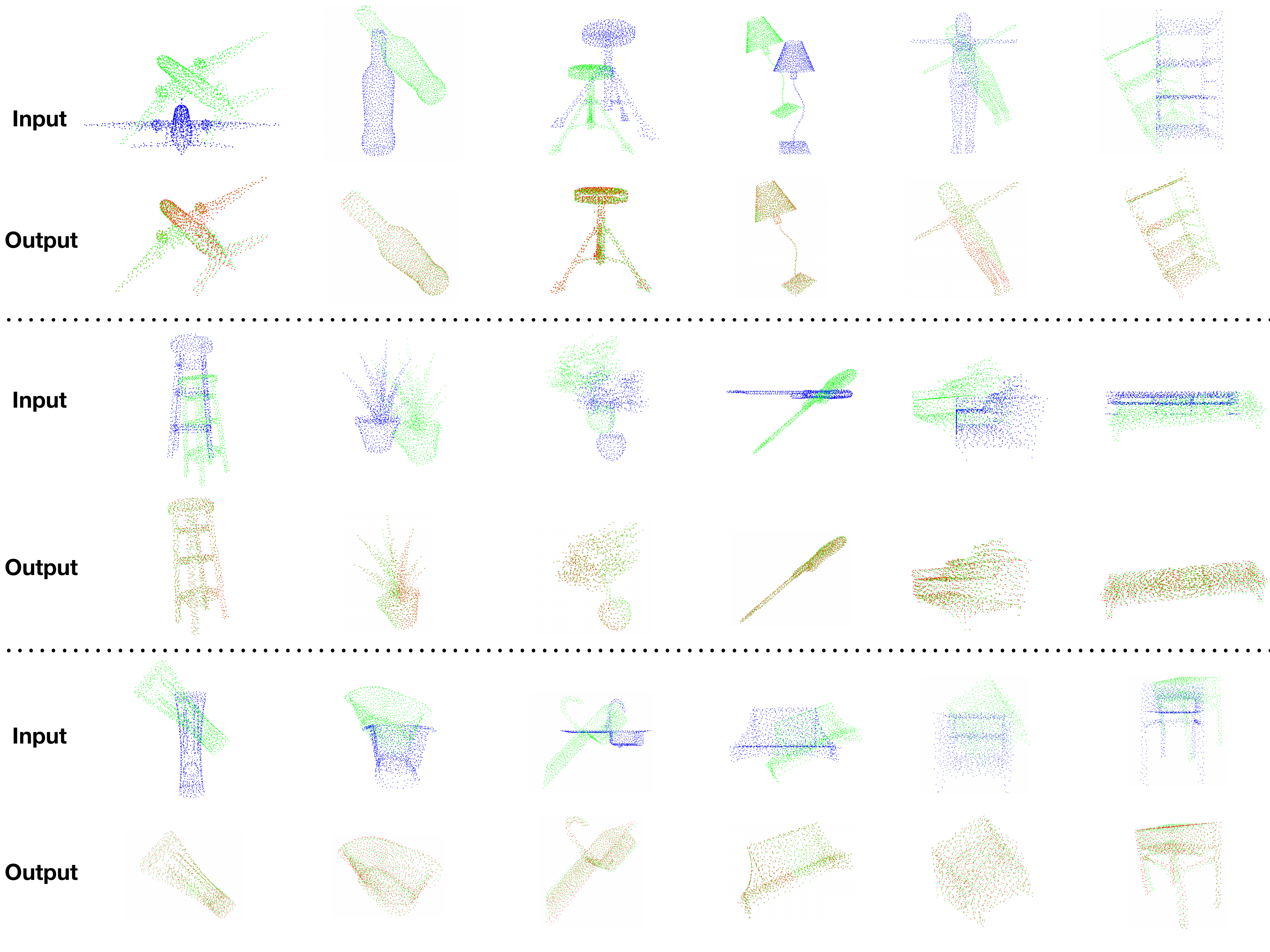}
  \caption{Results of DCP-v2. \textbf{Top}: inputs. \textbf{Bottom}: outputs of DCP-v2.  \label{fig:dcp-v2}}
\vskip 0.1in
\end{figure*}

Table \ref{table:full:modelnet40} evaluates performance of our method and its peers in this experiment (vanilla ICP nearly fails). DCP-v1 already outperforms other methods under all the performance metrics, and DCP-v2 exhibits even stronger performance. Figure \ref{fig:dcp-v2} shows results of DCP-v2 on some objects.  

\begin{table}[t] 
\vskip 0.1in
\begin{center}
\resizebox{\linewidth}{!}{
\begin{tabular}{lccccccr}
\toprule
Model & MSE($\bR$) & RMSE($\bR$) & MAE($\bR$)  & MSE($\bt$) & RMSE($\bt$) & MAE($\bt$)\\ 
\midrule
ICP &  894.897339 & 29.914835 & 23.544817 & 0.084643 & 0.290935 & 0.248755 \\
Go-ICP \cite{Yang2016GoICP} & 140.477325 & 11.852313 & 2.588463 & 0.000659 & 0.025665 & 0.007092 \\
FGR \cite{Zhou2016fgr} &  87.661491 & 9.362772 & 1.999290 & 0.000194 & 0.013939 & 0.002839 \\
PointNetLK \cite{Goforth2019} &   227.870331 & 15.095374 &  4.225304 & 0.000487 & 0.022065 & 0.005404 \\
\midrule
DCP-v1 (ours)  &  6.480572 & 2.545697 & 1.505548 & \textbf{0.000003} & \textbf{0.001763} & \textbf{0.001451} \\
DCP-v2 (ours) & \textbf{1.307329} & \textbf{1.143385} & \textbf{0.770573} & \textbf{0.000003} & 0.001786 & \textbf{0.001195} \\
\bottomrule
\end{tabular}
}
\end{center}
\caption{ModelNet40:  Test on unseen point clouds\label{table:full:modelnet40}}
\vskip 0.1in
\end{table}

\subsection{ModelNet40:  Category Split}
\label{sec:unseen:modelnet40}

To test the generalizability of different models, we split ModelNet40 evenly by category into training and testing sets. We train DCP and PointNetLK on the first 20 categories, then test them on the held-out categories. ICP, Go-ICP and FGR are also tested on the held-out categories. As shown in Table \ref{table:unseen:modelnet40}, on unseen categories, FGR behaves more strongly than other methods. DCP-v1 has much worse performance than DCP-v2, supporting our use of the attention module. Although the learned representations are task-dependent, DCP-v2 exhibits smaller error than others except FGR, including the learning-based method PointNetLK.  

\begin{table}[t]
\vskip 0.1in
\begin{center}
\resizebox{\linewidth}{!}{
\begin{tabular}{lcccccccccr}
\toprule
Model & MSE($\bR$) & RMSE($\bR$) & MAE($\bR$)  & MSE($\bt$) & RMSE($\bt$) & MAE($\bt$)\\ 
\midrule
ICP &  892.601135 & 29.876431 & 23.626110 & 0.086005 & 0.293266 & 0.251916 \\
Go-ICP \cite{Yang2016GoICP} & 192.258636 & 13.865736 & 2.914169 & 0.000491 & 0.022154 & 0.006219 \\
FGR \cite{Zhou2016fgr} &  97.002747 & 9.848997 & \textbf{1.445460} & 0.000182 & 0.013503 & 0.002231 \\
PointNetLK \cite{Goforth2019} &  306.323975 & 17.502113 & 5.280545 & 0.000784 & 0.028007 & 0.007203 \\
\midrule
DCP-v1 (ours) &  19.201385 & 4.381938 & 2.680408 & \textbf{0.000025} & \textbf{0.004950} & \textbf{0.003597} \\
DCP-v2 (ours) &  \textbf{9.923701} & \textbf{3.150191} &  2.007210 & \textbf{0.000025} & 0.005039 & 0.003703 \\
\bottomrule
\end{tabular}
}
\end{center}
\caption{ModelNet40: Test on unseen categories\label{table:unseen:modelnet40}}
\vskip 0.1in
\end{table}

\subsection{ModelNet40:  Resilience to Noise}
\label{sec:gaussian:modelnet}

We also experiment with adding noise to each point of the input point clouds. We sample noise independently from $\N(0, 0.01)$, clip the noise to $[-0.05, 0.05]$, 
and add it to $\X$ during testing. In this experiment, we use the model from \S\ref{sec:full:modelnet40} trained on noise-free data from all of ModelNet40.

Table \ref{table:gaussian:modelnet40} shows the results of this experiment. ICP typically converges to a far-away fixed point, and FGR is sensitive to noise. Go-ICP, PointNetLK and DCP, however, remain robust to noise.  

\begin{table}[t] 
\vskip 0.1in
\begin{center}
\resizebox{\linewidth}{!}{
\begin{tabular}{lcccccccccr}
\toprule
Model & MSE($\bR$) & RMSE($\bR$) & MAE($\bR$)  & MSE($\bt$) & RMSE($\bt$) & MAE($\bt$)\\ 
\midrule
ICP &  882.564209 & 29.707983 & 23.557217 & 0.084537 & 0.290752 & 0.249092 \\
Go-ICP \cite{Yang2016GoICP} & 131.182495 & 11.453493 & 2.534873 & 0.000531 & 0.023051 & 0.004192 \\
FGR \cite{Zhou2016fgr} & 607.694885 & 24.651468 & 10.055918 & 0.011876 & 0.108977 & 0.027393 \\
PointNetLK \cite{Goforth2019} &   256.155548 & 16.004860 & 4.595617 & 0.000465 & 0.021558 & 0.005652 \\
\midrule
DCP-v1 (ours) &  6.926589 & 2.631841 & 1.515879 & 0.000003 & 0.001801 & 0.001697 \\
DCP-v2 (ours) &  \textbf{1.169384}  & \textbf{1.081380} & \textbf{0.737479}  & \textbf{0.000002} & \textbf{0.001500} & \textbf{0.001053} \\
\bottomrule
\end{tabular}
}
\end{center}
\caption{ModelNet40: Test on objects with Gaussian noise \label{table:gaussian:modelnet40}}
\vskip 0.1in
\end{table}

\subsection{DCP Followed By ICP}

Since our experiments involve point clouds whose initial poses are far from aligned, ICP fails nearly every experiment we have presented so far. In large part, this failure is due to the lack of a good initial guess. As an alternative, we can use ICP as a \emph{local} algorithm by initializing ICP with a rigid transformation output from our DCP model. Figure \ref{fig:bimba} shows an example of this two-step procedure; while ICP fails at the global alignment task, with better initialization provided by DCP, it converges to the global optimum.  In some sense, this experiment shows how ICP can be an effective way to ``polish'' the alignment generated by DCP.

\subsection{Efficiency}
We profile the inference time of different methods on a desktop computer with an Intel I7-7700 CPU, an Nvidia GTX 1070 GPU, and 32G memory. Computational time is measured in seconds and is computed by averaging 100 results. As shown in Table \ref{table:efficiency}, DCP-v1 is the fastest method among our points of comparison, and DCP-v2 is only slower than vanilla ICP.

\begin{table}[t] 
\vskip 0.1in
\begin{center}
\resizebox{\linewidth}{!}{
\begin{tabular}{lccccccr}
\toprule
\# points & ICP & Go-ICP & FGR  & PointNetLK & DCP-v1 & DCP-v2\\ 
\midrule
512 & 0.003972 & 15.012375 & 0.033297 & 0.043228 & 0.003197 & 0.007932 \\
1024 & 0.004683 & 15.405995 & 0.088199 & 0.055630 & 0.003300 & 0.008295 \\
2048 & 0.044634 & 15.766001 & 0.138076 & 0.146121  & 0.040397 & 0.073697\\
4096 & 0.044585 & 15.984596 & 0.157124 & 0.162007 & 0.039984 & 0.74263\\
\bottomrule
\end{tabular}
}
\end{center}
\caption{Inference time (in seconds)\label{table:efficiency}}
\vskip 0.1in
\end{table}
\section{Ablation Study}
\vskip 0.1in

We conduct several ablation experiments in this section, dissecting DCP and replacing each part with an alternative to understand the value of our construction. 
All experiments are done in the same setting as the experiments in \S\ref{sec:full:modelnet40}.

\subsection{PointNet or DGCNN?}
\label{sec:ablation:PNorDGCNN}

We first try to answer whether the localized features gathered by DGCNN provide value over the coarser features that can be measured using the simpler PointNet model. As discussed in \cite{dgcnn}, PointNet \cite{QiSMG17} learns a global descriptor of the whole shape while DGCNN \cite{dgcnn} learns local geometric features via constructing the $k$-NN graph. We replace the DGCNN with PointNet (denoted as PN) and conduct the experiments in \S\ref{sec:full:modelnet40} on ModelNet40 \cite{Wu2015modelnet}, using DCP-v1 and DCP-v2. Table \ref{table:ablation:PNorDGCNN}. Models perform consistently better with DGCNN that their counterparts with PointNet.

\begin{table}[t] 
\vskip 0.1in
\begin{center}
\resizebox{\columnwidth}{!}{
\begin{tabular}{lccccr}
\toprule
Metrics & PN+DCP-v1, & DGCNN+DCP-v1  & PN+DCP-v2 & DGCNN+DCP-v2 \\
\midrule
MSE($\bR$) & 17.008427 & 6.480572 & 49.863022 & 1.307329 \\
RMSE($\bR$) & 4.124127 & 2.545697 & 7.061375 & 1.143385 \\
MAE($\bR$)  & 2.800184 & 1.505548 & 4.485052 & 0.770573 \\
MSE($\bt$) & 0.000697 & 0.000003 & 0.000258 & 0.000003 \\
RMSE($\bt$) & 0.026409 & 0.001763 & 0.016051 & 0.001786 \\
MAE($\bt$) & 0.01327 & 0.001451 & 0.010546 & 0.001195 \\ 
\bottomrule
\end{tabular}
}
\end{center}
\caption{Ablation study:  PointNet or DGCNN?\label{table:ablation:PNorDGCNN}}
\vskip 0.1in
\end{table}
 
\subsection{MLP or SVD?}

While MLP is in principle a universal approximator, our SVD layer is designed to compute a rigid motion specifically. In this experiment, we examine whether an MLP or a custom-designed layer is better for registration. We compare MLP and SVD with both DCP-v1 and DCP-v2 on ModelNet40. Table \ref{table:ablation:SVDorMLP} shows both DCP-v1 and DCP-v2 perform better with SVD layer than MLP. This supports our motivation to compute rigid transformation using SVD. 

\begin{table}[t] 
\vskip 0.1in
\begin{center}
\resizebox{\columnwidth}{!}{
\begin{tabular}{lccccr}
\toprule
Metrics & DCP-v1+MLP & DCP-v1+SVD  & DCP-v2+MLP & DCP-v2+SVD  \\
\midrule
MSE($\bR$) & 21.115917 & 6.480572 & 9.923701 & 1.307329 \\
RMSE($\bR$) & 4.595206 & 2.545697 & 3.150191 & 1.143385 \\
MAE($\bR$)  & 3.291298 & 1.505548 & 2.007210 & 0.770573 \\
MSE($\bt$) & 0.000861 & 0.000003 & 0.000025 & 0.000003 \\
RMSE($\bt$) & 0.029343 & 0.001763 & 0.005039 & 0.001786 \\
MAE($\bt$) & 0.022501 & 0.001451 & 0.003703 & 0.001195 \\ 
\bottomrule
\end{tabular}
}
\end{center}
\caption{Ablation study:  MLP or SVD?\label{table:ablation:SVDorMLP}}
\vskip 0.1in
\end{table}

\subsection{Embedding Dimension}

\cite{QiSMG17} remarks that the embedding dimension is an important parameter affecting the accuracy of point cloud deep learning models up to a critical threshold, after which there is an insignificant difference. To verify our choice of dimensionality, we compare models with embeddings into spaces of different dimensions. We test models with DCP-v1 and v2, using DGCNN to embed the point clouds into $\R^{512}$ or $\R^{1024}$. The results in Table \ref{table:ablation:moredims} show that increasing the embedding dimension from 512 to 1024 does marginally help DCP-v2, but for DCP-v1 there is small degeneracy. Our results are consistent with the hypothesis in \cite{QiSMG17}.

\begin{table}[t] 
\vskip 0.1in
\begin{center}
\resizebox{\columnwidth}{!}{
\begin{tabular}{lccccr}
\toprule
Metrics & DCP-v1 (512) & DCP-v1 (1024) & DCP-v2 (512) & DCP-v2 (1024) \\
\midrule
MSE($\bR$) & 6.480572 & 7.291216 & 1.307329 & 1.217545 \\
RMSE($\bR$) & 2.545697 & 2.700225 & 1.143385 & 1.103424 \\
MAE($\bR$)  & 1.505548 & 1.616465 & 0.770573 & 0.750242 \\
MSE($\bt$) & 0.000003 & 0.000001 & 0.000003 & 0.000003 \\
RMSE($\bt$) & 0.001763 & 0.001150 & 0.001786 & 0.001696 \\
MAE($\bt$) & 0.001451 & 0.000677 & 0.001195 & 0.001170 \\ 
\bottomrule
\end{tabular}
}
\end{center}
\caption{Ablation study: Embedding dimension\label{table:ablation:moredims}}
\vskip 0.1in
\end{table}

\section{Conclusion}



In some sense, the key observation in our Deep Closest Point technique is that learned features greatly facilitate rigid alignment algorithms; by incorporating DGCNN \cite{dgcnn} and an attention module, our model reliably extracts the correspondences needed to find rigid motions aligning two input point clouds.  Our end-to-end trainable model is reliable enough to extract a high-quality alignment in a single pass, which can be improved by iteration or ``polishing'' via classical ICP.

DCP is immediately applicable to rigid alignment problems as a drop-in replacement for ICP with improved behavior.  Beyond its direct usage, our experiments suggest several avenues for future inquiry.  One straightforward extension is to see if our learned embeddings transfer to other tasks like classification and segmentation.  We could also train DCP to be applied \emph{iteratively} (or recursively) to refine the alignment, rather than attempting to align in a single pass; insight from reinforcement learning could help refine approaches in this direction, using mean squared error as reward to learn a policy that controls when to stop iterating.  Finally, we can incorporate our method into larger pipelines to enable high-accuracy Simultaneous Localization and Mapping (SLAM) or Structure from Motion (SFM).
\section{Acknowledgement}
The authors acknowledge the generous support of Army Research Office grant W911NF-12-R-0011, of National Science Foundation grant IIS-1838071, from an Amazon Research Award, from the MIT-IBM Watson AI Laboratory, from the Toyota-CSAIL Joint Research Center, and from the Skoltech-MIT Next Generation Program. Yue Wang wants to thank David Palmer for helpful discussion. 

{\small
\bibliographystyle{ieee}
\bibliography{egbib}
}

\end{document}